# A Rubric for Human-like Agents and NeuroAI

## Ida Momennejad*

*Microsoft Research NYC*
*Reinforcement Learning Station*
*300 Lafayette, New York, New York,10012*
*ORCID: 0000-0003-0830-3973*

*Author for correspondence (idamo@microsoft.com).
†Present address: Reinforcement Learning Station, Microsoft Research NYC, 300 Lafayette, New York, New York, 10012, USA.

# Abstract

Researchers across cognitive, neuro-, and computer sciences increasingly reference "human-like" artificial intelligence and "neuroAI". However, the scope and use of the terms are often inconsistent. Contributed research ranges widely from mimicking *behaviour,* to testing machine learning methods as *neurally plausible* hypotheses at the cellular or functional levels, or solving *engineering* problems. However, it cannot be assumed nor expected that progress on one of these three goals will automatically translate to progress in others. Here a simple rubric is proposed to clarify the scope of individual contributions, grounded in their commitments to *human-like behaviour, neural plausibility*, or *benchmark/engineering* goals. This is clarified using examples of weak and strong neuroAI and human-like agents, and discussing the generative, corroborate, and corrective ways in which the three dimensions interact with one another. The author maintains that future progress in artificial intelligence will need strong interactions across the disciplines, with iterative feedback loops and meticulous validity tests–leading to both known and yet-unknown advances that may span decades to come.

What do we mean by human-like AI? What makes an algorithm, representation, or architecture neurally plausible or brain-like– be it the brains of drosophila, rats, or humans? Researchers across and within cognitive, neuro-, and computer sciences increasingly vary in what they mean when they use terms such as human-like AI [1–3], and its close relatives, human-inspired, human-level [4], bio-plausible [5], animal AI [6], as well as brain-inspired [7], neurally plausible, neuroscience-inspired [8], or more recently neuroAI [6,8–15]. This paper proposes that a closer look at research in recent decades reveals three main dimensions of goals and contributions across these fields: a commitment to achieving agents with *human-like* (or animal-like) *behaviour*, *neural plausibility*, or to solving specific *computer science or engineering* problems (mostly irrelevant to neuroscience but may still be the goal of brain-like solutions).

Why is it important to distinguish these goals and contributions? An implicit but frequent assumption across these fields is that progress in one goal will automatically and readily translate to progress in others, and that at some point this cumulative progress will somehow stumble upon artificial general intelligence, 'AGI' [16]. However, in practice not only are these dimensions not guaranteed to translate to one another, but insistence on conflating the contributions of these approaches may slow down, stall, or block progress.

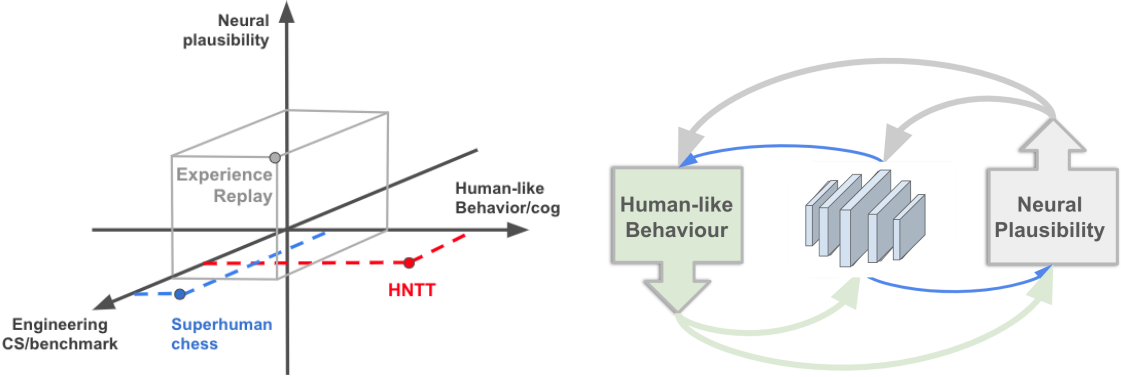

**Figure 1.The Rubric.** *(Left)* The rubric of goals and contributions toward human-like agents and neuroAI captures three dimensions of goals and contributions: human-like behaviour, neural plausibility, and engineering/computer science. While work at the intersection of the three is common, success in one dimension does not guarantee success in another. Examples are marked, e.g., research on experience replay has contributed to all dimensions (see Section 2), while the Human Navigation Turing Test (HNTT) has mostly contributed to engineering goals and human studies with no strong neuro contribution (Section 3), and superhuman chess AI is an engineering contribution but does not produce human-like action sequences and did not contribute to neuro research (Section 3). *(Right)* Progress in AI, human-like agents, and neuroAI requires interactions and feedback loops among all dimensions that can be *generative* of novel research, or *corroborative* or *corrective* of existing work.

Thus, in order to clarify the scope and commitments of individual and historically connected contributions to the field, here a simple rubric is proposed with the three dimensions of *human-like behaviour [1,17,18]*, *neural plausibility [9]*, and *engineering* [19,20] to assess which goals a given study prioritises and to what extent (Figure 1). Note that since the rubric is not binary, it allows us to assess the intersectional contributions of studies and their bridging as well. The scale is meant to be qualitative, it suffices to say that a given study, for example, may contribute to the goal of building agents with human-like behaviour, with only a low to moderate commitment to neural plausibility, and no contribution to engineering goals.

It is the author's hope that by the end of this paper, the reader appreciates the following key points of the rubric. First, success in one dimension does not guarantee success in others. For instance, the author believes that computer science research with no human-like or neuroAI focus is not likely to accidentally stumble upon a machine learning elixir that mimics and explains both brain and behaviour. Second, true neurally plausible research requires a direction of influence from neuroscience to AI as well. This is not simply satisfied by importing machine learning approaches as hypotheses of neural circuits or as data analysis methods, or assuming all neural networks are neurally plausible given their brain-inspired ancestry in the 1940s or 1980s. Third, the dimensions proposed here are not exhaustive and can be expanded to cover varieties of goals and scales (e.g., single neuron vs. large scale brain signals) in contemporary psychology and neuroscience [21], embodied approaches to behaviour [22–24], and robotics [25]. In the interest of space, most examples will focus on human-oriented studies, missing the wealth of animal-oriented neuroAI research that has been the focus of computational and systems neuroscience [6,9,21,26,27]. Finally, progress in building more intelligent AI, human-like agents, and neuroAI requires well-thought commitments and interactions among the dimensions, rather than merely importing successful methods in one to another or focusing on a handful of famous methods (see Discussion).

*Why these three dimensions?* Historically, artificial intelligence and machine learning are rooted in the imitation of human behaviour and behavioural studies of animal learning (e.g., trial and error learning)–even prior to neuroscience inspiration [28–31]. In his 1950 article, the first to seriously probe the question of AI, Turing suggests that human behaviour is the guide we must use to develop artificial intelligence [29]. Thus, comparison to human behaviour has been a defining feature of AI from its very inception, finding its appropriate place back in AI research in the past decades thanks to datasets, benchmarks, and user studies Neuroscience has inspired much of the architectural evolution of early artificial intelligence, especially neural networks [28–32], and many neuroscientists emphasise the importance of behaviour as a key mode of neuroscience research [33,34]. However, in spite of this rich history, more recently neuroscience to AI influence has been more rare, and particularly neglected–especially given 'scale' and ever-larger models and stronger corporate incentives for solving engineering goals have risen to fashion.

Given these historical ebbs and flows of the connections among the three fields, recently the term *NeuroAI* has been used as an umbrella term for research that is crucially committed to *bidirectional* interactions between AI and neuroscience [6,8–15,35], and *human-like or human-level agents* are built with contributions from all fields [1,4]. *NeuroAI* covers a wide spectrum: from AI that is weakly inspired by neuroscience to agents with close correspondence to the brain's algorithms, architectures, or representations.

Moreover, while a relationship between neuroAI and behaviour is assumed, AI's focus on human-like behaviour has been mostly related to *predicting* behaviour for personalisation, human-computer interaction, or consumer research, while the cognitive sciences attempt to *explain* and *understand* behaviour as a scientific discipline. In other words, while these three dimensions have historically interacted, a lack of their individuation may lead to the loss of progress–or funding–in crucial academic fields. Note that these three dimensions are by no means mutually exclusive, binary, or exhaustive, and the claim here is not that all AI researchers should care about neuroAI or the human-likeness of their algorithms. The rubric's goal is to simply provide orientation to varying goals, contributions, and intersections, and clarify how their interactions can advance the fields [1,2,36].

*Generative, corroborative, and corrective interactions.* Each dimension can contribute to others in *generative*, *corroborative*, and *corrective* ways. For instance, successful models of human and animal behaviour are often later used as hypotheses to predict or explain the brain's underlying representations and algorithms [17,37–43]. As such, neuroAI can *corroborate* the validity of successful *behavioural* models, helping researchers choose which model is more *neurally plausible*. This, in turn, can lead to designing novel behavioural tasks to distinguish among similarly neurally plausible algorithms. Such a feedback loop among dimensions advances all (Figure 1). In the other direction, AI approaches that capture aspects of neural algorithms, architecture, or representations (e.g., convolutional neural networks that capture hierarchical vision in inferior temporal cortex, see section 1) are sometimes used to predict human-like behaviour and detailed correspondence with the brain [44,45].

For instance, upon the observation of behavioural mismatch between a neurally plausible algorithm and human behaviour (e.g., CNNs' poor performance in the face of adversarial images and patches), some neuroAI researchers bring back the focus to stronger human-like behaviour (e.g., testing adversarial examples for humans, noting that strong neuroAI needs to show similar errors in the face of adversarial examples to humans), leading to *revisions* and *corrections* in the original models [46]. In what follows we will discuss recent successful examples of such intersectional research, from perception and navigation to associative learning, multitasking, reinforcement learning, memory, and planning.

Section 1 discusses neurally plausible AI. This dimension's primary goal is building models with neurally plausible algorithms, representations, or specific functions; or contributing data and benchmarks to test the neural plausibility of existing AI models. Section 2 discusses research on agents with human-like behaviour. This dimension's primary concern is building computational agents that match the nuances of human behaviour and experiments that test them. Section 3 discusses research that is primarily focused on engineering, benchmarks, and other computer science goals. Whether through building deep learning algorithms, robotics, or novel benchmarks; the computer science and engineering goal is to make theoretical advances, solve theoretical or real-world problems, or adapt AI for specific applications. Section 4 discusses future directions. The aim of the rubric proposed here is to orient the reader to the interactions of the dimensions, with examples from a variety of human-like agents, neuroAI, and engineering contributions, from models of the visual cortex to Turing tests for 3D navigation in Xbox games and single- and multi-agent reinforcement learning.

## 1. What We Talk about when We Talk About NeuroAI

This Section first defines neurally plausible AI and neuroAI. We then define weak and strong neuroAI with examples, with a focus on neuroAI domains that historically have iterative interactions with other dimensions in generative, corroborative, and corrective ways. The author hopes that the examples in the rubric, as well as discussion of challenges and opportunities, can help better clarify the contributions of any given approach and easier navigate potential future directions.

What do we mean by *neuroAI* [15], *neuroscience inspired AI* [8], and *neurally plausible* AI? In light of this paper's aims, these terms fall on a spectrum of approaches with either or multiple of the following goals. *Building* AI agents whose algorithm, representations, or functions are neurally plausible; *applying* AI or behavioural models as hypotheses in neuroscience (effectively testing or *corroborating* them for neural plausibility) or analysis tools for neural data; *generating* datasets, tasks, and benchmarks that benefit engineering or behavioural research; or offering evidence to update or *correct* models in other dimensions.

About a decade ago neural network approaches to AI moved the field from a winter to a spring. Since then, a body of work has focused on various relationships between machine learning and AI, writ large, and neuroscience. The wide spectrum of long-term visions ranged from the use of deep neural networks in computational neuroscience [27,47], biological attention [48], memory [49,50], navigation [51,52], the mutual interactions between neuroscience and ML [11], neuroscience-inspired AI [8], drawing analogues across mental imagery and deep learning [53], learning to learn or meta-learning as prefrontal cortex (PFC) theory [54], followed by considerations of fast and slow reinforcement learning [55], continual learning [56], human-centred robotics [25], and critical perspectives [10]. While some researchers predominantly focused on importing and testing deep RL methods or large language models as hypotheses for mechanisms in the brain [44,54,57], others brought the focus to bidirectional interaction of the fields, especially to how neuroscience influences AI [9].

*Weak and Strong NeuroAI.* The author proposes that research committed to neural plausibility can, but is not guaranteed to, contribute to other dimensions in *generative, corroborative, or corrective* ways, and that depending on the level of commitment and contributions, neuroAI research can fall on a spectrum from *weak* to *strong neuroAI*. While *weak neuroAI* loosely applies tools from other disciplines to solve problems primarily in one discipline [8,58], here *strong neuroAI* is defined as neurally plausible approaches with *strong* bidirectional or three-way goals and contributions across the dimensions.

*Weak neuroAI* research may simply take out of the box ML algorithms to either model or analyse neural data without a strong claim that the brain uses a closely similar algorithms to generate it [59], as well as AI that employs methods with vaguely neuroscience-inspired ancestry but has no commitment to novel neurally plausible contributions. Both weak and strong NeuroAI could pursue the *application* or *corroboration* of existing contributions, e.g., testing convolutional neural networks (CNN) as a falsifiable hypothesis of the brain's mechanisms or representations [44,60], or using transformers, as in language models that improve linguistic skills with Generative Pre-Training (GPT) [61], as hypotheses of the similarity structure in language-related fMRI representations [57].

The spectrum towards *strong neuroAI* goals and contributions include incrementally stronger claims in both AI to neuroscience and neuroscience to AI directions, of which the neuroscience to AI influence is particularly signifying [9,62]. *Strong neuroAI* can integrate deep learning with neuroscience [27], use deep learning as hypotheses for neural representations [46,51,57], or explicitly build or test brain-like algorithms [35] and architectures. *Strong* neuroAI contributes to novel behavioural predictions or new engineering solutions. Examples include incorporating neurally plausible neurotransmitters or travelling waves in deep learning solutions [63,64], generating appropriate data or benchmarks, or neuromorphic computing [65], in which hardware is designed after brain circuits to empower efficient computation. Finally, strong NeuroAI could offer *corrections* to existing models, such as building a hippocampal-like architecture for navigation, e.g., RatSLAM [66], or integrating bio-plausible synaptic learning rules, separate excitatory and inhibitory units, and segregated dendrites [67].

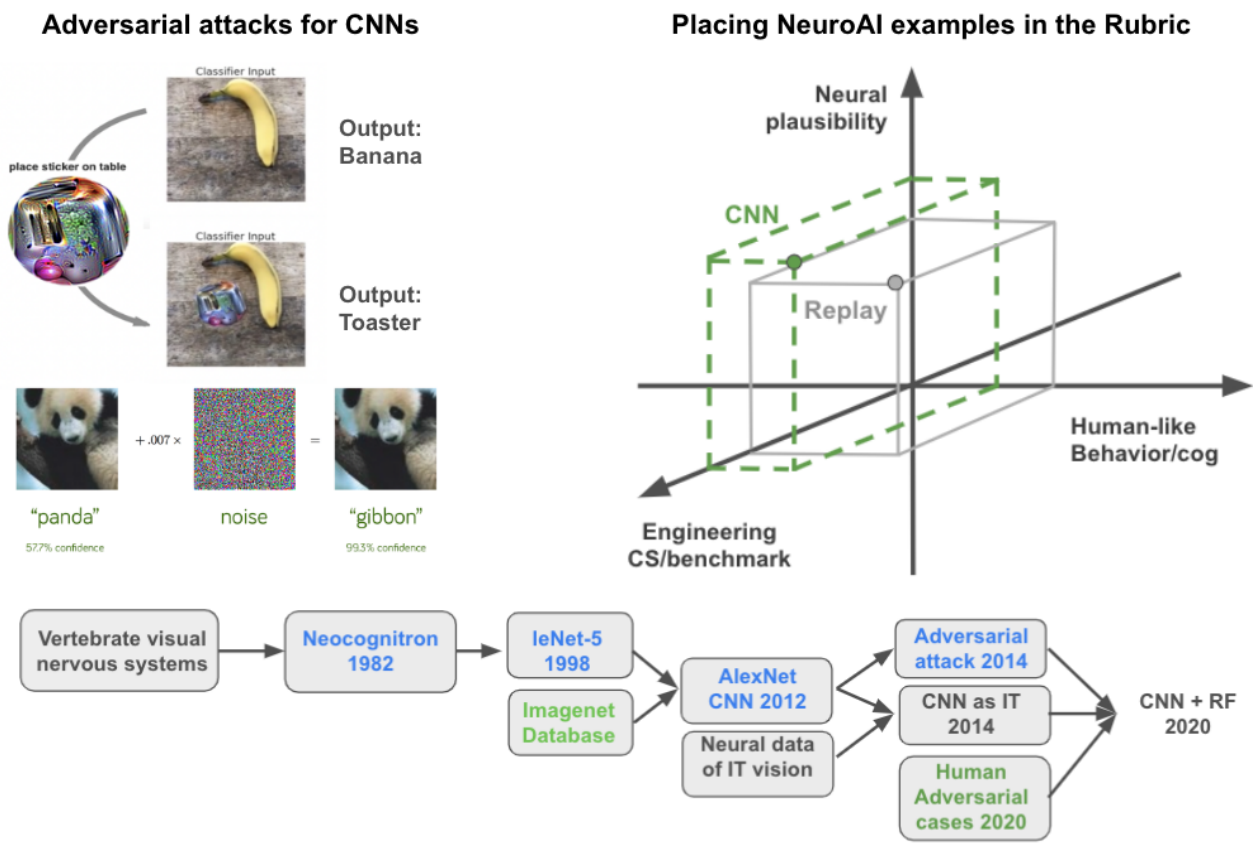

**Figure 2. NeuroAI examples in the rubric.** (*Top Left*) Convolutional neural networks (CNNs) have been proposed as models of the primate ventral visual cortex [44]. However, they are not robust to adversarial images that humans are robust to [68][69]. Conversely, adversarially trained deep learning models display superhuman behaviour, not making the same errors as humans. In both cases behaviour is not human-like. (*Top Right*) NeuroAI examples in the rubric. Note how the extent of their contribution to each dimension differs. (Bottom) An example of historical generative, corroborate, and corrective interactions across the three dimensions (*green font: human behaviour, black font: neurally plausible, blue font: computer science; CNN + RF: Convolutional neural networks plus non-uniform sampling (like the primate retina) and a range of receptive field sizes*).

*NeuroAI examples in the rubric, challenges, and opportunities.* Successful ML approaches such as convolutional neural networks have historical neuroscience inspiration: four decades ago, Fukushima's Neocognitron was an explicit attempt to model the simple and complex cells observed in the visual cortex by Hubel & Weisel [70], (Figure 2, bottom), and it is the grandparent of modern CNNs, including Lecun's LeNet in 1998 and AlexNet in 2012, decades later [32,71,72]. Since 2014, a series of elegant neuroAI studies have compared object and image recognition in convolutional neural networks (CNNs)–and other deep and recurrent neural networks of vision–[73] with neural data of monkey and human vision [44]. The hypothesis was that the hierarchy of representations in CNN layers captures the hierarchical processing and representations in the inferotemporal cortex (IT) for object recognition. CNN's top hidden layers were used to predict inferior temporal cortical responses in spiking neurons and populations.

This approach was met with great enthusiasm, heralding a wave of papers testing deep learning techniques as models of neural activity [47,60,74–77], inspiring novel ideas in the philosophy of neuroscience [78,79]. However, these approaches were not robust to adversarial attacks, i.e., misclassify images with an amount of added noise imperceptible to humans [68,80] (Figure 2). The models also failed in the presence of adversarial patches that were merely placed next to the object but completely changed the classification [69] (Figure 2).

Following these challenges, adversarially trained and generative neural networks rose to popularity [81] and unsupervised models of IT vision were proposed [82]. However, these adversarially-trained or generative approaches sometimes display superhuman behaviour and are not tested for a match to human-like perception [83]. The next advances came from collecting behavioural data on human perception of adversarial images, and the addition of biologically inspired approaches to adversarial robustness such as non-uniform sampling (like the primate retina) and a range of receptive field sizes [84], transfer [85], and innovative approaches to improving the alignment between models and behaviour [46,86,87]. However, no algorithm to date has been shown to match human-level perception in terms of both successes and failures in the face of adversarial images. This is partly due to lack of comprehensive *behavioural* datasets regarding human performance on varieties of adversarial images. Thus, not only did the feedback loop among dimensions generate this field, but behavioural studies of adversarial images for humans may hold the key to improving architectures to match human behaviour (i.e., human successes, errors, and reaction times) as well as neural responses.

Another example of algorithms used as hypotheses of brain function is research using the reinforcement learning (RL) computational framework. RL is a machine learning framework in which an agent is rewarded for choosing optimal sequences of actions that maximise reward in a given environment. While this framework was mostly inspired by early twentieth century psychology of trial and error behaviour, as well as later theories of associative learning and neural computation [88,89], recent advances in RL capture learning cognitive maps, generalisation and transfer, metalearning, and learning structures and representations [90]. Some RL algorithms are tested as hypotheses of neural representations underpinning underlying human behaviour [54,90], behaviourally successful RL algorithms are tested as hypotheses on neuroimaging data [38,43,91], novel RL agents are proposed to test and refine AI agents in a feedback loop with human experimentation [2], AI agents are tested for flexible behaviour using environments inspired by human experiments [36], or multi-agent deep RL is designed to show comparable hierarchical innovation in problem solving to humans [92].

Since RL is a large area of research, let us focus on one successful example from this field, experience replay [93], as another case of interactions across the different fields over the past decades. The term was first coined by a computer scientist in 1992 [94] at Carnegie Mellon and was adopted within a few years in a seminal cognitive neuroscience paper [95]. It has since been a staple of focus of numerous human and animal behavioural, computational, and neuroscience studies as well as engineering solutions [90,96–100]. Notably, deep reinforcement learning agents began to show progress on the Atari suite following the use of experience replay [4], and both AI and human- and neuro-oriented replay-based approaches improved with more nuanced replay models [101–104]. Thus, experience replay offers a good example of an approach contributing to, and borrowing from, all dimensions of the rubric (Figure 2).

In this section we discussed weak and strong neurally plausible agents and neuroAI, with generative, corroborate, and corrective contributions to other dimensions. We also discussed neuroAI examples with interactions among the three dimensions of the rubric. More contributions from other dimensions to and interactions with the neural plausibility dimension are discussed in the following sections, and more future directions are discussed in the final section.

## 2. What We Talk About When We Talk About Human-like Behaviour

This section regards how measuring and comparing *behaviour in human and artificial agents* contributes to all dimensions. In some examples research on human-like agents may import ML models as hypotheses to *corroborate*; *correct* existing agents using behaviour or benchmarks; or *generate* new agents, datasets, and benchmarks toward more human-like intelligent behaviour and cognition. We will also note the limits of tasks for which an agent can show human-like behaviour, and wrap up with rubric examples of human behaviour in interaction with AI and neuroscience. The key proposal is that experiments and models of human behaviour can contribute to building agents with intelligent behaviour and neural plausibility, e.g., by testing the behavioural flexibility of their models (leading to *corroboration* or *correction* of those agents).

The interaction of this dimension with other dimensions can offer standards for task design, human-level benchmarks, and standards of model-comparison to ensure consistency and quality of research.

*What motivates building artificial agents with human-like behaviour?* Besides researchers who believe human intelligence is the only real guide towards general artificial intelligence (AGI) [105], motivations and goals for building human-like agents range from scientific to practical. They can range from understanding or predicting human behaviour and inferring cognitive and neural processes from behaviour [17,33,34], to computational psychiatry [106–108], and to building agents with human-like play in an Xbox game [1] or AI that interacts with humans in practical applications, e.g., for building better HCI (Human Computer Interaction) and BCI (Brain Computer Interface) [20], simulating interventions, AI for healthcare, or other purposes [19].

Varieties of research on human-like agents may focus on studying *animal behaviour* as a stepping stone to better understanding evolutionary roots or specificity of human behaviour, e.g., rat navigation, [6]; on the use of Turing Tests, e.g., the Navigation Turing Test (NTT) for 3D video games [1–3]; or on deep learning and other machine learning approaches to analysing or generating behaviour (e.g., for intuitive physics [109]). Behavioural research may also focus on *evolutionary* models, e.g., investigating evolutionary sources of behaviour [22–24,110] or varieties of *multi-agent* problem solving [92,111,112]. Experiments inspired by human behaviour can also be used to test the flexibility and generalizability of existing AI agents [36].

*What does it take for an artificial agent to show human-like behaviour?* Some may suggest that meeting or exceeding an overall human-level benchmark in video games, chess, or broadly overall scores on specific tasks are sufficient. However, this 'overall score' view fails at capturing human-like errors and latency of responses, as well as the *flexibility* and *transfer* of the same behaviour given minor changes in the task or environment–let alone generalisation to different contexts. Here we define research on *human-like agents* with either one or multiple of the following goals. i) Building *computational models and agents that match the nuances of human behaviour*; ii) designing and conducting *behavioural experiments to test agent behaviour for flexibility or compare it with that of humans* (or other animals, though here we focus on human behaviour).

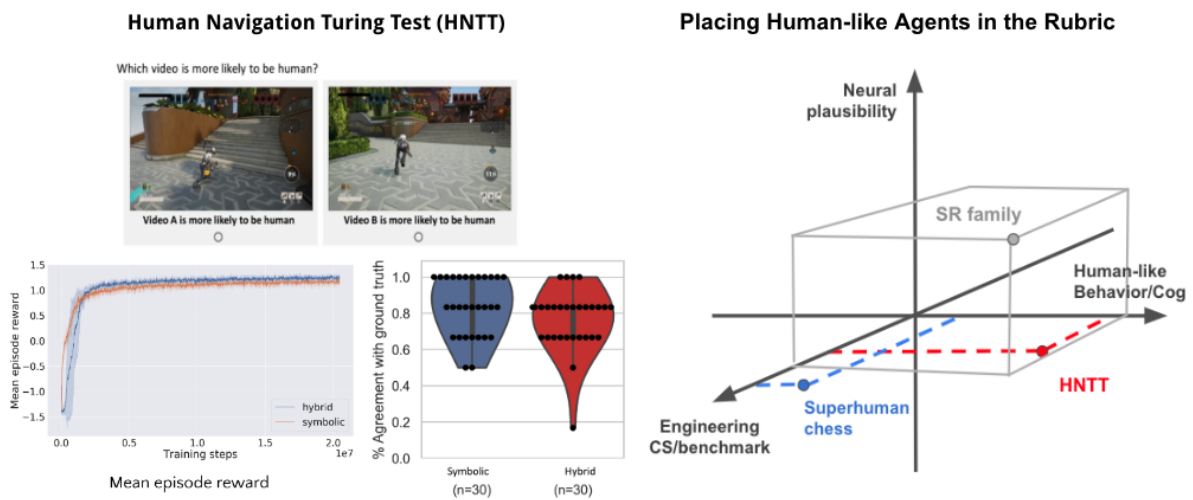

**Figure 3. Example human-like agents in the rubric.** *(Left)* A Turing test for human-like Navigation was designed [1] to compare the performance of two recent state of the art deep learning agents for navigation (Symbolic vs. Hybrid architectures)[113]. Although both agents passed benchmarks with similar mean rewards over episode, neither agents passed the Turing Test, and the Hybrid agent was judged more human-like in direct comparison. *(Right)* HNTT and other human-like agents in the rubric. Superhuman chess AI contributes to AI but does not produce human-like action sequences nor do they contribute to neuro research, while a family of models related to successor representation have contributed to all three.

*How do we measure and compare behaviour? Behaviour* can be operationalized and measured in terms of *accuracy* and *errors* of responses, trajectory of moves, e.g., in video games or a chess game [3], or differences in *learning rates* and *reaction times* across conditions or tasks [114], e.g., in memory or value-based decision-making, or based on human judgments, i.e., as in Turing Tests [1]. Cognitive science, psychology, and cognitive neuroscience experiments typically infer internal cognitive and mental states using behavioural measures as well. Pairing seemingly simple behavioural measurements with strong theory or computational models, it is possible to infer cognitive maps [115], algorithms, or representations used to solve a task [17,18,42,116]. This, in turn, can lead to *generating* novel neural or computational hypotheses, *corroborating* existing ones, or providing *corrections* to existing neural or computational theories of cognition.

Regardless of the specific behavioural measure used, the author proposes the terms *strong* and *weak human-like behaviour* to distinguish contributions to this dimension as follows. *Strong human-like behaviour* identifies agents that display the full array of measured behaviour (accuracies, errors, reaction times, differences in across conditions) on a given task or set of tasks. Such agents or models are expected to also support accurate generalisation and transfer to other related tasks or contexts using the same parameters. *Weak human-like behaviour* refers to agents that match a single overall score for a game or task or set of tasks, e.g., overall Atari scores or chess scores, but are neither tested for nor required to guarantee matching behavioural nuances e.g., superhuman chess AI does not play chess with human-like action sequences [3]–primarily because such programs were primarily motivated by the engineering axis.

*Placing examples in the rubric.* It is important to highlight research on human behaviour that contributes to other dimensions in generative, corroborative, or corrective ways.

*Human Navigation Turing Test.* The first example used human experiments to improve the human-like navigation of AI agents in video games. To this end, two recent papers introduced the Human Navigation Turing Test (HNTT) for a 3D video game (Bleeding Edge on XBox, Figure 3). The HNTT was designed to test whether state of the art (SoTA) AI for navigation in 3D video games, with similar benchmark scores, behaved human-like to human observers [1]. In HNTT, observers were shown pairs of videos of gameplay and asked to judge which one was more likely played by an average human player rather than AI. While some trials included a human vs. agent video, unbeknownst to the participant judges, in some conditions the videos were played by different artificial agents (with similar benchmark scores) making it possible to directly compare the human-like performance of the two SoTA navigation agents (Figure 2).

The study [1] and its replication [2] revealed that the two SoTA agents [113], with the same performance on previous navigation benchmarks, were not judged equally human-like by human judges, and neither passed the navigation Turing test (Figure 3). In addition, six artificial agents with different architectures and training measures were designed to judge the human-likeness of videos as well, a test called Artificial NTT (ANTT). Almost all artificial judges could predict which video was played by a human with above chance accuracy when the ground truth included a human player. However, compared to human subject's ratings of two agent videos, all artificial judges reliably failed at judging which of two artificial agents looked more human-like to humans. This research reveals the importance of human experimental research on the perception of human-like behaviour across agents and offers opportunities for future artificial judges.

*Superhuman chess.* The second example comes from an illuminating study, which investigated whether AI with superhuman chess performance produced human-like sequences during the game. The authors analysed the extent to which the agent's chess strategies and sequences of moves were recognized as "human-like" by expert chess players, and found that they were not [3]. These results suggest that the challenge of playing *superhuman* chess and the challenge of playing *human-like* chess are quite different (Figure 3), a fact that in spite of seemingly being obvious has been overlooked by the field. Now some may argue that humans could improve their chess performance if they learn the superhuman play from the AI. However, just as some physical ranges of motion are unnatural for the human body, certain algorithmically generated moves may never be human-like either–as the authors observed in unpublished follow up research, teaching the AI generated sequences to humans did not improve their game. Finally, while there are excellent examples of robotics and embodiment research that align with human-like or neuroAI goals [117,118], either being *inspired by* it or *inspiring* it, engineering goals do not require that they be aligned.

*SR family.* The next example regards a series of behavioural and neuroscience studies on a family of learners that store predictive representations of successive states (the *successor representations or SR [119]*). Models and experiments on this *SR family* of agents capture behavioural and neural responses in multi-scale planning, navigation, and associative learning in humans and rodents [17,38,43] and contribute to engineering solutions with deep successor feature learning methods [41,120,121]. The successor representation and its eigenvectors also capture place fields and grid fields in rodent hippocampal during navigation [37,39,42,90]. In recent years, a number of innovative approaches complement this cumulative evidence [40,51,52,64,122–126].

This cohesive body of work, which has largely been produced in a close conversation among the scientists across disciplines, displays a well-functioning example of progress achieved through iterative feedback in the study of algorithms, representations, neuroAI, and behaviour in multiple species, showing converging and complementary results from multiple computational approaches across global labs. Doing this body of work justice would require its own separate paper with dedicated scope [90]. However, the author hopes to have conveyed the collaboration and feedback across neuroAI, engineering, and behaviour across studies in these examples, captured in the rubric as '*SR family*' (Figure 3).

*Inference.* Another behavioural contribution is noteworthy (somewhat related to the SR family but not shown in the figure due to space constraints). A long-standing puzzle in the field of cognitive and computational neuroscience concerns the nature of multi-step inference in the hippocampus, which requires stitching together separately learned associations. A simplified example would be to learn AB and BC associative pairs and *infer* AC. In the past decade, two opposing computational hypotheses were proposed for the mechanism underlying this inference. One model suggests that one-step associations and recurrence underlie the brain's inference [127], while the other, complementary learning systems within the hippocampus, suggests that the brain learns the multi-step association as well, making inference readily available [128]. Interestingly, a behavioural study was specifically designed to distinguish the predictions of the recurrent vs. the multi-step representation models [129]. The study shows that interleaved presentation of the AB and BC pairs assist the formation of the AC inference. This finding is not in line with the recurrence model, which predicts no difference in the order of stimulus presentation.

Similarly, a set of successor representation studies designed experiments to compare different reinforcement learning agents with human behaviour, and showed that it is not one-step associations unrolled at decision time but learned and cached multi-step representations that capture human behaviour in retrospective revaluation, or transfer of, rewards and transition structure [17]. In this paradigm participants learned two sequences, ABC and XYZ, in phase 1. In phase 2, they were shown only the pairs BZ and YC, signalling that the transition structure of the initial sequence had changed. In phase 3, participants were asked to make value-based decisions that relied on making correct inferences about the A to Z and X to C multi-step contingencies. It was shown that the successor representations updated via offline replay, SR-Dyna, outperformed model-based RL (which stores 1-step transitions) in capturing these findings [17,130]. In line with the inference example above, this behavioural result supports multi-step representations as well. Two different fMRI studies then offered evidence of neural plausibility for these findings using multi-scale SR in realistic virtual reality navigation of city distances [43] and comparing the inference and SR methods in a statistical learning task [131].

*Challenges and opportunities.* It is common to simply import successful AI models as hypotheses of human or animal behaviour. For instance, consider reinforcement learning (RL), a computational framework where an agent takes actions in an environment to achieve certain goals or maximise rewards [30]. Existing RL models and algorithms are often used readily as models of human behaviour or brain function, or modified to match behaviour or neural data [17]. The interaction of RL and psychology/neuroAI has been prolific given RL's historical inspirations from psychology and neuroscience [63,132], and contemporary RL's ability to learn the structure of the environment or to plan, replay, and generalise– all of which are common in humans and animals [90]. Over the past decade, the marriage of RL with deep neural networks has brought on an AI spring of sorts [4,13] that inspired much of the literature reviewed here [8].

Often once RL or deep RL agents display successful behaviour in complex tasks, such as Atari games, their behaviour is compared with humans on other tasks. First, candidate agents are selected, then specific tasks [133] are designed to compare the behaviour of the agents with human–or animal– performance on the same task [17,18,42]. This takes place in the form of matching differences in accuracies, errors, and reaction times–either in a more qualitative manner or more precisely. The outcome is typically the rejection of all but one or two of the candidate models, which can then be subjected to neuroscientific hypothesis testing for corroboration [38,43,64,122]. At its best, this iterative process also leads to new algorithms inspired by the results that are more aligned with behaviour and brains–in terms of the underlying algorithms, architectures, or representations.

Other research uses deep neural networks as models for human behaviour [134], or use deep learning to study human-like intuitive physics [109] and multi-agent problem solving [92]. However, so far the field has not defined checks and standards to ensure adequate similarity between behavioural comparisons, experimental tasks, and comparable environments designed. This absence of standards for task design, benchmarks, and model comparison in the field may challenge progress. While a number of papers and libraries attempt to provide such standard frameworks to address these challenges [135,136], including a recent library for neurally plausible learning and navigation, Neuro-Nav [133], much more work as well as critical post-publication reviews are needed to ensure progress in this area.

*No free lunch: Behaviour on how many tasks?* The "no free lunch theorem" suggests we cannot expect any agent to generalise to all tasks [137]. Thus, another key challenge is the set of tasks (set A) an agent needs to show human-like behaviour on to generalise to a reasonable set of other tasks (set B). It is important to build robust algorithms that do not fail at learning novel tasks (B), nor forget tasks learned earlier (A) due to catastrophic forgetting. While multitasking and continual RL have substantially contributed to this direction, many challenges and opportunities remain for future work [56,138–141].

In this section we have discussed examples of the interaction between agents with human-like behaviour and other dimensions. First, human experiments can inspire benchmarks or experiments to test and *corroborate* the behavioural flexibility of agents. Second, human judgments can compare, *correct*, and improve the human-like behaviour of AI agents. Third, experiments, datasets, and benchmarks of human behaviour can help *generate* novel neuroscience or engineering models and experiments. More standards for task design as well as behavioural and model comparison are required to ensure sustainable consistency and quality of research in the field.

## 3. What We Mean by Computer Science/Engineering Goals

This section regards research that is primarily focused on engineering, benchmarks, and computer science goals. Whether through building deep learning algorithms, robotics, or novel benchmarks; computer science and engineering goals include *making theoretical and empirical advances in computer science, solving theoretical or real-world problems, and adapting AI for specific applications*. In fact, the majority of AI research is applied to diverse domains, including scientific problems [142], most of which have nothing to do with biological behaviour or neuroscience (just as much of computer science is not about AI).

Sometimes artificial agents built with an engineering goal in mind may happen to match or exceed human overall scores, or task-specific benchmark measures, e.g., the overall score in Atari games. However, that is not *necessarily* always their primary goal nor contribution (though it could be). Moreover, it has also been shown that even AI agents with the same overall benchmark scores are not necessarily judged to be equally human-like by human judges. In fact, benchmarks may contribute in early stages of model development. For instance, consider Brain-Score [85], a platform for comparing computational models of vision against brain benchmarks using neural measurements of the ventral IT cortex. Brain-Score results show that up to a certain point better performance on image recognition correlated with better matches to neural and behavioural data. However, once the agent's behaviour gets closer to being human-like, the agent's performance on benchmarks cannot predict which model is likely to be judged as more human-like, e.g., in chess [1–3], nor in 3D navigation in video games as in the example of HNTT in section 2 [1–3](Figure 3).

Previous sections discussed a few cases of what the author would like to call *benchmark chasing and its discontents*. For instance, vision models with high performance on brain benchmarks did not necessarily show human-like behaviour in the face of adversarial images [68], and navigation agents with high scores on human benchmarks did not show human-like navigation behaviour in video games [1]. In what follows we will also discuss deep MBRL agents with high scores on Atari benchmarks that, when tested on tasks for testing human model-based behaviour, did not show the flexible behaviour expected of model-based agents with a model of the environment [36]. In spite of the focus of the present paper, the author acknowledges that not all AI research is, nor needs to be, dedicated to connections between neural and cognitive sciences. That said, it is the author's hope that the following examples offer motivations for AI researchers with engineering goals to look to the cognitive and neuro- sciences for inspiration (also see discussion for potential future directions).

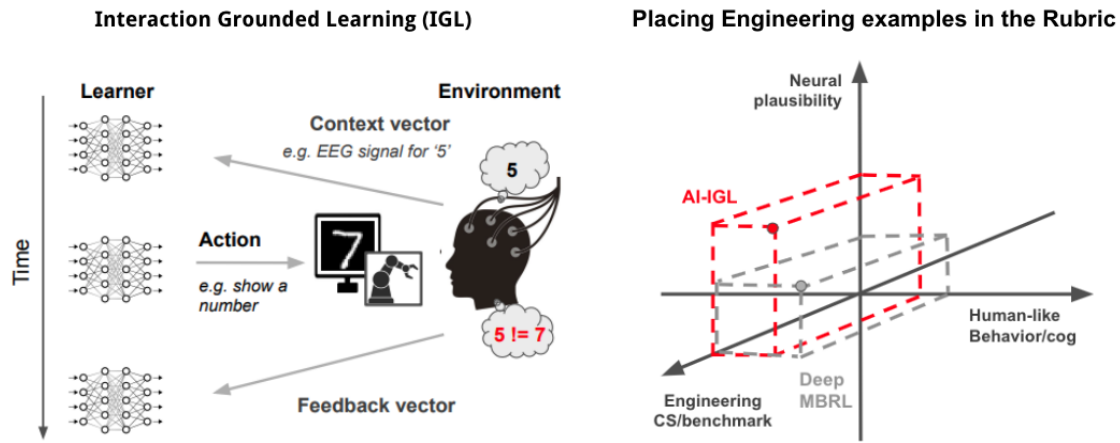

**Figure 4. Computer science examples in the rubric.** *(Left)* Interaction Grounded Learning (IGL) is a computer science solution to the problem of a device directly interacting with biological signals, and in the absence of interpretable signals or any grounding of what those signals mean [19,20]. *(Right)* IGL and another engineering example in the rubric. A recent paper, inspired by experiments in human cognitive neuroscience, designed simple transfer tasks (similar to reward revaluation [17,18,100]) to test the flexibility of existing deep model-based RL agents (DreamerV2, PlaNet, muZero). The results revealed that neither deep MBRL agent passed the flexibility test that model-based agents are expected to pass, but changing replay in these agents could solve this challenge [36].

*Examples in the rubric.* Interaction Grounded Learning (IGL) or Action Inclusive IGL is placed in the rubric (Figure 4). This is an example of a computer science solution to the problem of a device directly interacting with biological signals (e.g., interacting with brain signals as in brain computer interface (BCI) or interacting with muscles and nerves as prosthetics) in the absence of supervision or grounding of what the signals mean [19,20].

The second example involves recent research, which draws inspiration from *human experiments* to test the adaptivity of deep model based algorithms [36,143]. Researchers wanted to test whether popular deep MBRL agents, i.e., muZero [144], PlaNet [145], DreamerV2 [146], show the behavioural flexibility expected of model-based agents, i.e., RL agents that can flexibly change their action policies using an internal model of the environment that is stored independently of rewards. Since well-known human experiments had already been designed to test model-based flexible behaviour in humans, the researchers adapted a similar experimental setting and task to test the behaviour of MBRL agents [17,18].

The task was simply that the agents learned the path to a reward location, then observed partial changes in the environment that indicated that the rewards had moved. Finally, the agents were tested on whether they could integrate this information to flexibly adapt their trajectory toward the new reward location. Theoretically, all model-based RL (MBRL) algorithms are expected to have the flexibility needed to perform such tasks. Surprisingly, the authors found that in spite of deep MBRL's impressive performance on the Atari suite benchmark, none passed this test of human-like model-based flexibility. This research also led to a novel algorithmic addition (using replay) that improved the behavioural flexibility of MBRL, offering a promising direction for behaviourally inspired testing and improving of mainstream AI.

Another example of successful interactions between the engineering and other dimensions (not shown in the rubric) is SAPIENS. This is a recent framework for creating multi-agent AI capable of human-like collective innovation. Specifically, the goal was hierarchical discovery in a "little alchemy" task where the agents had to combine elements to come up with new hierarchical discoveries, e.g., earth plus water gives mud is a level 1 discovery, fire plus mud gives brick is a level 2 discovery [92]. Human studies had shown that when all members of a social network communicated with one another they reached more similar shared memories [111,112]. However, when a network of humans faced hierarchical problem solving as in little alchemy, they benefited more from a dynamic graph, which alternated between agents working in a clustered network and periods of exchanging members across clusters [147,148].

Inspired by the network structures of these human studies, researchers connected multiple deep RL agents to solve the problem together either in a graph with all-to-all connections, or in a dynamic graph alternating between a clustered formation and all-to-all connections. The agents in the collective could share their experience with each other (similar to human communication) by sharing the content of their *replay* buffer. The structure and order in which these agents could share experience was predetermined according to different network topologies inspired by human studies. Consistent with human findings, the multi-agent networks of agents that shared experience in clusters first, and then exchanged members (dynamic topology) achieved the highest level of hierarchical innovation across tasks. In the future, the extension of such multi-agent studies, paired with meta learning [54], could offer a window into the evolution of human-like behaviour.

*Challenges and opportunities.* Whether solving Brain Computer Interface (BCI) challenges, or finance or robotics problems, engineering oriented researchers typically build agents for specific tasks or domains, e.g., pass or exceed human performance on benchmarks (such as performance on 57 Atari games). While sometimes engineering goals can intersect or diverge with the other dimensions, models with equal benchmark performance are not necessarily scored as equally human-like (see Navigation Turing Tests judged earlier). In other words, a challenge of overall benchmark measures is that while they may give the impression of human-level AI, they are often not well poised to capture the nuances of human-like behaviour, might be game-able via shortcut learning or spurious correlations, and are in need of interactions with the other dimensions.

Another major challenge for benchmark chasing and behaviour matching is *generalisation*. While it is unreasonable to expect any agent to generalise to all tasks, as postulated by the "no free lunch theorem" discussed earlier [137], it is reasonable to compare agents with regards to their capacity for generalisation. The challenge here is that different algorithms with more or less similar human-like behaviour on a given task (e.g., categorising cats and dogs) could radically differ with regards to how well they generalise to other tasks–or how likely their underlying mechanisms, representations, and algorithms are to be neurally plausible. Moreover, while outperforming human performance on a benchmark is desired for the engineering goal, exceeding human benchmarks is not aligned with the goals of human-like AI (Section 2).

*Are large language models examples of strong interactions across the dimensions?* The past few years mark the rise of large language models (LLMs). It is remarkable that a simple objective of predicting the next word, given a complex enough architecture and sufficiently large text corpus, has yielded such impressive behaviour [149]. Similar to testing CNNs as models of IT cortex, recent neuroAI research has investigated similarities between LLMs and the brain's language system [57].

While a detailed summary is outside the scope here, a challenge is that LLMs are known to be inefficient architectures with billions of parameters. Pruning studies have shown it is possible to prune 90% or more parameters from trained neural networks without hurting their ability [150], and the lottery ticket hypothesis suggests that it's possible to prune those parameters before or early in training without harming the network's ability to learn [151]. Nonetheless, most industry-led LLM approaches have not been focused on pruning solutions. It is the author's opinion that meticulous behavioural and neurally plausible considerations (e.g., learning from pruning in the brain) may prove necessary, if not unavoidable, for making LLM architectures more efficient–rather than relying on ever-scaling. Scaling LLMs even larger can pose environmental challenges and neglects much opportunity for growth [152].

Also note that LLMs are trained on large corpora of human-generated text and are capable of pulling out what is statistically probable as the next word, or chunks of words. However, humans use language to express inner worlds or communicative needs, intent, or understanding within their perception of the environment's context, whereas LLMs are simply *trained on human output*, not grounded in the conditions, needs, and processes that generate it. To remedy this, grounded language learning research uses a simulated reality, in which the agent's language is grounded, rather than merely parroting the content of human-produced linguistic corpora [153,154]. The aim of this research is to recover the truth values of perceptual statements by counterfactual simulations of a virtual world. However, this approach does not sufficiently address all challenges raised above. Careful interdisciplinary work is needed in this area.

In the author's opinion, the likelihood of ever larger LLMs stumbling upon precise brain-like architectures and representations, which can generalise similarly to humans and pass meticulous tests of human-likeness, is slim. Unfortunately, there is also a risk that insisting on testing this hypothesis may well bankrupt the planet's already exhausted resources [155]. Even if by the will of science fiction gods such an LLM emerges by sheer scaling, the precise analysis and understanding of the relevant architectures would require a parallel neuroscience of its own [156].

While it is possible to imagine a neuroscience of LLMs as a subfield of neuroAI, the author warns against falling prey to–or being led astray by– what can be called the "all you need" bubble: the call of some ML or neuroAI papers for undue investment in one approach at the expense of others, backed by a cottage industry of sensational titles, debates, and social media posts that inflate how a single approach will help us stumble upon artificial general intelligence, or even sentience, e.g., "scale is all you need", "reward is enough", etc. [157].

Even if such an ML elixir existed, neuroAI aficionados would still need an understanding of various brain functions; a single canonical computation principle or circuit cannot address all research topics in all dimensions. That being said, irrespective of such promises, it is unlikely that any single approach will yield a holy grail theory of everything related to intelligence and brains. The goal of understanding the brain's architectures, functions, and mechanisms with the various tools of neuroAI and human-like behaviour will carry on outside the bubble, interacting with and benefiting from the bubble's byproducts.

## 4. Discussion and Future Directions

A simple rubric is proposed here to capture goals and contributions to research in human-like AI, neuroAI, and CS/engineering (Figures 1-4). We have discussed how research in these dimensions can lead to *generating*, *corroborating or applying*, and even *correcting* achievements of other dimensions–in *weak* and *strong* ways. The author does not claim that these dimensions are exhaustive. The hope is simply that this rubric helps to critically orient the reader in an ever-expanding literature at the intersection of cognitive science, neuroscience, and AI, without considering any specific approach as the "*all you need*" elixir of general intelligence or ignoring what lies outside one's primary dimension. Recognizing this interconnected space of goals and contributions is a step toward more effective bridging and progress through feedback loops and collaborations across dimensions.

Hopefully by now the reader agrees that benchmark chasing and superhuman AI for specific tasks, while very valid goals for the engineering or intersecting goals, do not capture all dimensions in the rubric proposed here–but can serve as a step in decades long exchanges across dimensions (see Section 1 and Figure 2). As an example, some suggest that generic models that leverage computation will outperform domain-specific models that require human knowledge, with the example of deep search as the key to computer chess defeating Kasparov in 1997 [158]. However, this view merely aligns with CS/engineering goals of simply solving problems, and as discussed in Section 2, recent research shows that superhuman chess AI does not create action sequences that are recognized as human-made to chess experts [3](Figure 3).

The author maintains that engineering approaches by themselves are not likely to address all the commitments and goals of human-like AI and neuroAI, just as research in each of those dimensions alone is not guaranteed to lead to progress in others. Researchers motivated by biological behaviour and neuroAI are urged to critically and meticulously examine claims and methods they import from the computer science and engineering areas rather than accepting them out of the box. On the other hand, given the computational and energy *inefficiency* of current large scale AI approaches, drawing inspiration from the flexibility of prefrontal architectures and the behavioural adaptivity of ecological intelligence (be it of human or octopus) can largely benefit the engineering-oriented research toward artificial intelligence.

*Challenges and future opportunities*. In recent years major AI conferences encourage engineers and computer scientists to draw inspirations from, and discuss contributions to, real world problems as well as scientific research. While some have been fruitful to the field of neuroAI [58,159], a few challenges and opportunities in these multidisciplinary interactions are noteworthy.

Some engineering and computer science papers oversimplify notions imported from psychology and neurosciences and make exaggerated and scientifically ungrounded claims of having discovered 'all you need' to achieve them or use them as a solution to age old problems. Examples are plenty, from attention (which relates to selectively attending to external stimuli in psychology but is closer to indexed memory in transformer models [160]) to working memory and sentience. Critical inquiry and exchange among the dimensions offer *opportunities* to counter inflated claims and contribute to a more careful and sustainable interdisciplinary science of intelligence. To this end, the hope is to see *stronger neuroAI* research with algorithms, representations, and architectures that predict both the biological *behaviour* in question and align with its *neural underpinnings* (see examples in Sections 1 and 2).

Can we identify a set of algorithms and rules that can be combined to achieve human-like and neurally plausible mechanisms? What level of brain-likeness would we stop at? Would we stop at large networks of various cognitive functions or demand the further brain-like realisation of cellular architecture, layers, and circuits? To paraphrase Lewis Carroll's most cited quote in cognitive map research, we cannot use a human-made one-to-one map of reality as it would cover all of physical space [161]. That said, setting up standards and protocols of what counts as brain-like is not arbitrary and requires multi-level diversity of contributions.

*Opportunities: Pruning, PFC inspired-, and ecological AI.* Other than ecological intelligence, among the remarkable capacities of biological brains is their energy efficiency–even though the focus is often on AGI-related skills such as generalisation, abstraction, transfer, sample efficiency, etc. As pruning studies on AI show, current deep learning models often have wasteful architectures, since after pruning the neural network, research finds that the entire large architecture can be made much more compressed [162–164]. There are different possible ways to improve the efficiency of architectures. One is creating more flexible architectures with better capacity for generalisation and abstraction [165,166], planning, and tracking hierarchical prospective tasks [167] adaptive to the demands of the environment. Since the prefrontal cortex (PFC)–especially the anterior PFC– is the epitome of these traits, it is the author's belief that a key neuroAI progress moving forward will come from AI with PFC-inspired architectures [54,167–169].

The author maintains that the human PFC can learn, host, and switch among many abstract algorithms and representations upon demand and dependent on the context and task demands. The PFC enables metalearning [54], and is responsive to neuromodulators (such as norepinephrine or dopamine) that signal changes in the environment, helping optimise hyperparameters or change gears from one algorithm to another [170]. This is thanks to its widespread interconnectivity within the PFC, as well as bidirectional promiscuous connections to the brains' hubs (i.e., regions with high connectivity) and earlier sensory areas in the "back of the brain". From empowering abstract task sets [171], goal-directed and adaptive behaviour [172], and long-term and prospective future plans [167,173]; to compositional abstractions and basis sets, generative basis sets, and hierarchical [174,175] and analogical reasoning [176], more meticulous research on PFC algorithms, representations, and architectures can inspire more efficient AI architectures. This may also herald a more energy efficient era of architectures and agents, by achieving these traits in a hypothesis driven manner as opposed to leaving it all to the hands of scaling and fate.

Another emerging direction is that of evolutionary and *ecologically inspired AI*. It has been suggested that studying major synthetic evolutionary transitions and their universal traits [177], as well as tracing the phylogenetic development of learning across evolutionary scales [24], can inspire the development of AI. Moreover, human behavioural ecology suggests that intelligence can only be as complex as its environment requires, proposing an environment-oriented approach to understanding intelligence [110]. Along the same lines, an Animal AI test bed [178] has been proposed, Animal AI Olympics, which is inspired by the evolution of self-control across species [22], offering a promising first path toward AI with ecologically valid behaviour.

Together, a new subspace of progress within the rubric is carved by architectures better equipped with the ability for generalisation and transfer, including PFC-like architectures, and more human-like (or octopus-like) ecological adaptivity for learning to want what they need–even in sparse reward contexts [179]. Such progress may help solve deep learning's radical energy inefficiency, challenges with generalisation and transfer, sample inefficiency, and the labour and planetary costs of extracting resources required to scale AI further [155]. Taking it a step further, perhaps groundbreaking future progress will come from "*wilding*" computation, a term the author proposes as a shorthand for building architectures that are built of *biomatter* rather than silicon, e.g., mycelial networks. The path to such wild innovation may well be to step in the dark and outside our existing preconceptions of artificial intelligence, armed with state of the art knowledge across dimensions, to open up new avenues of scientific investigation.

Finally, in spite of historical and notable examples of interaction across the dimensions discussed above, the majority of today's everyday AI research progresses *without* drawing direct inspirations from neuroscience– and even less often does it strive to be neurally plausible. It may be tempting to expand the notion of *neural plausibility* or *neuroAI* to include any ML or AI study using a neural network with loose archeological *neuro-inspired* lineage multiple decades removed. However, the author maintains that such loose criteria for neuroAI may promote a sort of laziness in drawing helpful strong neuroscience inspirations, falling short of actualizing the many novel and productive ways in which neuroscience *can* contribute to improving modern AI/ML. That said, it is noteworthy that sometimes approaches that were originally weakly inspired by neuroscience are later used as strong hypotheses for brain function and mechanisms [44], while other times an algorithm, such as Simultaneous Localization and Mapping (SLAM) for navigation [180–182], evolved to span both robot navigation and the brain's algorithms for navigation behaviour, see RatSLAM [66] for rodent navigation. Thus, it seems important to highlight that interdisciplinary interactions of all kinds, weak and strong, across dimensions are likely to be fruitful in both known and unknown ways across long horizons of decades, benefiting all.

In short, the rubric of AI research proposed here specifically aims at highlighting commitments, contributions, and intersections of human-like agents, neuroAI, and engineering. The hope is that the proposed dimensions help the reader orient themselves in the space of every growing interdisciplinary AI research, and consider future directions that are friendly to the planet equipped with knowledge of ecological and biological intelligence, without losing sight of the bigger picture.

## Acknowledgments

The author is immensely grateful to Melanie Mitchell, John Krakauer, Arthur Juliani, and Blake Richards for comments; and Paul Linton and the two anonymous reviewers for significantly contributing to the overall structure, clarity, and quality of the paper. The author thanks conversations with Tony Zador, Anna Schapiro, MacKenzie Mathis, Eleni Nisioti, Clement Moulin-Frier, Katja Hoffman, Sam Devlin, Konrad Kording, Kate Crawford, Yoshua Bengio, Paul Cisek and other colleagues for inspiring various parts of the present paper over the years.

**Competing Interests**
*I have no competing interests.*

# References

1.  Devlin S, Georgescu R, Momennejad I, Rzepecki J, Zuniga E, Costello G, Leroy G, Shaw A, Hofmann K. 2021 Navigation Turing Test (NTT): Learning to Evaluate Human-Like Navigation. *ICML*

2.  Zuniga E *et al.* 2022 How Humans Perceive Human-like Behavior in Video Game Navigation. In *Extended Abstracts of the 2022 CHI Conference on Human Factors in Computing Systems*, pp. 1–11. New York, NY, USA: Association for Computing Machinery.

3.  McIlroy-Young R, Sen S, Kleinberg J, Anderson A. 2020 Aligning Superhuman AI with Human Behavior: Chess as a Model System. In *Proceedings of the 26th ACM SIGKDD International Conference on Knowledge Discovery & Data Mining*, pp. 1677–1687. New York, NY, USA: Association for Computing Machinery.

4.  Mnih V *et al.* 2015 Human-level control through deep reinforcement learning. *Nature* **518**, 529–533.

5.  Song Y, Lukasiewicz T, Xu Z, Bogacz R. 2020 Can the Brain Do Backpropagation? -Exact Implementation of Backpropagation in Predictive Coding Networks. *Adv. Neural Inf. Process. Syst.* **33**, 22566–22579.

6.  Zador AM. 2019 A critique of pure learning and what artificial neural networks can learn from animal brains. *Nat. Commun.* **10**, 3770.

7.  Poo M-M. 2018 Towards brain-inspired artificial intelligence. *Natl Sci Rev* **5**, 785–785.

8.  Hassabis D, Kumaran D, Summerfield C, Botvinick M. 2017 Neuroscience-Inspired Artificial Intelligence. *Neuron* **95**, 245–258.

9.  Richards BA *et al.* 2019 A deep learning framework for neuroscience. *Nat. Neurosci.* **22**, 1761–1770.

10. Saxe A, Nelli S, Summerfield C. 2020 If deep learning is the answer, then what is the question? *arXiv [q-bio.NC]*.

11. Storrs KR, Kriegeskorte N. 2019 Deep learning for cognitive neuroscience. *arXiv preprint arXiv:1903.01458*

12. Bertolero MA, Bassett DS. 2020 Deep Neural Networks Carve the Brain at its Joints. *arXiv [q-bio.NC]*.

13. Botvinick M, Wang JX, Dabney W, Miller KJ, Kurth-Nelson Z. 2020 Deep Reinforcement Learning and Its Neuroscientific Implications. *Neuron* **107**, 603–616.

14. Lake BM, Ullman TD, Tenenbaum JB, Gershman SJ. 2017 Building machines that learn and think like people. *Behav. Brain Sci.* **40**, e253.

15. Zador A *et al.* 2022 Toward Next-Generation Artificial Intelligence: Catalyzing the NeuroAI Revolution. *arXiv [cs.AI]*.

16. Goertzel B. 2014 Artificial General Intelligence: Concept, State of the Art, and Future Prospects. *Journal of Artificial General Intelligence* **5**, 1–48.

17. Momennejad I, Russek EM, Cheong JH, Botvinick MM, Daw ND, Gershman SJ. 2017 The successor representation in human reinforcement learning. *Nat Hum Behav* **1**, 680–692.

18. Daw ND, Gershman SJ, Seymour B, Dayan P, Dolan RJ. 2011 Model-Based Influences on Humans' Choices and Striatal Prediction Errors. *Neuron* **69**, 1204–1215.

19. Xie T, Langford J, Mineiro P, Momennejad I. 2021 Interaction-Grounded Learning. *arXiv [cs.LG]*.

20. Xie T, Saran A, Foster DJ, Molu L, Momennejad I, Jiang N, Mineiro P, Langford J. 2022 Interaction-Grounded Learning with Action-inclusive Feedback. *arXiv [cs.LG]*.

21. Kording KP, Blohm G, Schrater P, Kay K. 2020 Appreciating the variety of goals in computational neuroscience. *arXiv [q-bio.NC]*.

22. MacLean EL *et al.* 2014 The evolution of self-control. *Proc. Natl. Acad. Sci. U. S. A.* **111**, E2140–8.

23. Howard D, Eiben AE, Kennedy DF, Mouret J-B, Valencia P, Winkler D. 2019 Evolving embodied intelligence from materials to machines. *Nature Machine Intelligence* **1**, 12–19.

24. Cisek P. 2019 Resynthesizing behavior through phylogenetic refinement. *Atten. Percept. Psychophys.* **81**, 2265–2287.

25. Doncieux S, Chatila R, Straube S, Kirchner F. 2022 Human-centered AI and robotics. *AI Perspectives* **4**, 1–14.

26. Marder E, Taylor AL. 2011 Multiple models to capture the variability in biological neurons and networks. *Nat. Neurosci.* **14**, 133–138.

27. Marblestone AH, Wayne G, Kording KP. 2016 Toward an Integration of Deep Learning and Neuroscience. *Front. Comput. Neurosci.* **10**, 94.

28. McCulloch WS, Pitts W. 1943 A logical calculus of the ideas immanent in nervous activity. *The Bulletin of Mathematical Biophysics*. **5**, 115–133. (doi:10.1007/bf02478259)

29. Turing AM. 1950 I.—COMPUTING MACHINERY AND INTELLIGENCE. *Mind* **LIX**, 433–460.

30. Sutton RS, Barto AG. 2018 *Reinforcement Learning: An Introduction*. MIT Press.

31. Lindsay LG. 2021 *Models of the Mind: How Physics, Engineering and Mathematics Have Shaped Our Understanding of the Brain*. Bloomsbury Publishing (UK).

32. Fukushima K, Miyake S. 1982 Neocognitron: A Self-Organizing Neural Network Model for a Mechanism of Visual Pattern Recognition. In *Competition and Cooperation in Neural Nets*, pp. 267–285. Springer Berlin Heidelberg.

33. Krakauer JW, Ghazanfar AA, Gomez-Marin A, MacIver MA, Poeppel D. 2017 Neuroscience Needs Behavior: Correcting a Reductionist Bias. *Neuron* **93**, 480–490.

34. Niv Y. 2021 The primacy of behavioral research for understanding the brain. *Behav. Neurosci.* **135**, 601–609.

35. Nonaka S, Majima K, Aoki SC, Kamitani Y. 2021 Brain hierarchy score: Which deep neural networks are hierarchically brain-like? *iScience* **24**, 103013.

36. Wan Y, Rahimi-Kalahroudi A, Rajendran J, Momennejad I, Chandar S, van Seijen H. 2022 Towards Evaluating Adaptivity of Model-Based Reinforcement Learning Methods. *arXiv [cs.LG]*.

37. Stachenfeld KL, Botvinick MM, Gershman SJ. 2017 The hippocampus as a predictive map. *Nat. Neurosci.* **20**, 1643–1653.

38. Garvert MM, Dolan RJ, Behrens TE. 04 27, 2017 A map of abstract relational knowledge in the human hippocampal-entorhinal cortex. *Elife* **6**. (doi:10.7554/eLife.17086)

39. Momennejad I, Howard MW. 2018 Predicting the future with multi-scale successor representations. *BioRxiv*

40. de Cothi W, Barry C. 2020 Neurobiological successor features for spatial navigation. *Hippocampus* **30**, 1347–1355.

41. Vertes E, Sahani M. 2019 A neurally plausible model learns successor representations in partially observable environments. *arXiv [stat.ML]*.

42. Bellmund JLS, de Cothi W, Ruiter TA, Nau M, Barry C, Doeller CF. 2020 Deforming the metric of cognitive maps distorts memory. *Nat Hum Behav* **4**, 177–188.

43. Brunec IK, Momennejad I. 2021 Predictive Representations in Hippocampal and Prefrontal Hierarchies. *J. Neurosci.* (doi:10.1523/JNEUROSCI.1327-21.2021)

44. Yamins DLK, Hong H, Cadieu CF, Solomon EA, Seibert D, DiCarlo JJ. 2014 Performance-optimized hierarchical models predict neural responses in higher visual cortex. *Proc. Natl. Acad. Sci. U. S. A.* **111**, 8619–8624.

45. Higgins I, Chang L, Langston V, Hassabis D, Summerfield C, Tsao D, Botvinick M. 2021 Unsupervised deep learning identifies semantic disentanglement in single inferotemporal face patch neurons. *Nat. Commun.* **12**, 6456.

46. Dapello J, Kar K, Schrimpf M, Geary R, Ferguson M, Cox DD, DiCarlo JJ. 2022 Aligning Model and Macaque Inferior Temporal Cortex Representations Improves Model-to-Human Behavioral Alignment and Adversarial Robustness. *bioRxiv*. , 2022.07.01.498495. (doi:10.1101/2022.07.01.498495)

47. Kriegeskorte N. 2015 Deep Neural Networks: A New Framework for Modeling Biological Vision and Brain Information Processing. *Annu. Rev. Vis. Sci.* **1**, 417–446.

48. Lindsay GW, Miller KD. 2018 How biological attention mechanisms improve task performance in a large-scale visual system model. *Elife* **7**. (doi:10.7554/eLife.38105)

49. Banino A *et al.* 2020 MEMO: A Deep Network for Flexible Combination of Episodic Memories. *arXiv [cs.LG]*.

50. Pritzel A, Uria B, Srinivasan S, Badia AP, Vinyals O, Hassabis D, Wierstra D, Blundell C. 06--11 Aug 2017 Neural Episodic Control. In *Proceedings of the 34th International Conference on Machine Learning* (eds D Precup, YW Teh), pp. 2827–2836. PMLR.

51. Banino A *et al.* 2018 Vector-based navigation using grid-like representations in artificial agents. *Nature* (doi:10.1038/s41586-018-0102-6)

52. Whittington JCR, Muller TH, Mark S, Chen G, Barry C, Burgess N, Behrens TEJ. 2019 The Tolman-Eichenbaum Machine: Unifying space and relational memory through generalisation in the hippocampal formation. *bioRxiv* (doi:10.1101/770495)

53. Hamrick JB. 2019 Analogues of mental simulation and imagination in deep learning. *Current Opinion in Behavioral Sciences*

54. Wang JX, Kurth-Nelson Z, Kumaran D, Tirumala D, Soyer H, Leibo JZ, Hassabis D, Botvinick M. 2018 Prefrontal cortex as a meta-reinforcement learning system. *Nat. Neurosci.* **21**, 860–868.

55. Botvinick M, Ritter S, Wang JX, Kurth-Nelson Z, Blundell C, Hassabis D. 2019 Reinforcement Learning, Fast and Slow. *Trends Cogn. Sci.* **23**, 408–422.

56. Riemer M, Cases I, Ajemian R, Liu M, Rish I, Tu Y, Tesauro G. 2018 Learning to Learn without Forgetting by Maximizing Transfer and Minimizing Interference. *arXiv [cs.LG]*.

57. Schrimpf M, Blank IA, Tuckute G, Kauf C, Hosseini EA, Kanwisher N, Tenenbaum JB, Fedorenko E. 2021 The neural architecture of language: Integrative modeling converges on predictive processing. *Proceedings of the National Academy of Sciences* **118**, e2105646118.

58. Goyal A, Lamb A, Hoffmann J, Sodhani S. 2019 Recurrent independent mechanisms. *arXiv preprint arXiv*

59. Khosla M, Ngo GH, Jamison K, Kuceyeski A, Sabuncu MR. 2021 Cortical response to naturalistic stimuli is largely predictable with deep neural networks. *Sci Adv* **7**. (doi:10.1126/sciadv.abe7547)

60. Spoerer CJ, McClure P, Kriegeskorte N. 2017 Recurrent Convolutional Neural Networks: A Better Model of Biological Object Recognition. *Front. Psychol.* **8**, 1551.

61. Radford A, Narasimhan K. 2018 Improving Language Understanding by Generative Pre-Training.

62. Macpherson T, Churchland A, Sejnowski T, DiCarlo J, Kamitani Y, Takahashi H, Hikida T. 2021 Natural and Artificial Intelligence: A brief introduction to the interplay between AI and neuroscience research. *Neural Netw.* **144**, 603–613.

63. Dabney W, Kurth-Nelson Z, Uchida N, Starkweather CK, Hassabis D, Munos R, Botvinick M. 2020 A distributional code for value in dopamine-based reinforcement learning. *Nature* **577**, 671–675.

64. George TM, de Cothi W, Stachenfeld K, Barry C. 2022 Rapid learning of predictive maps with STDP and theta phase precession. *bioRxiv.* , 2022.04.20.488882. (doi:10.1101/2022.04.20.488882)

65. Roy K, Jaiswal A, Panda P. 2019 Towards spike-based machine intelligence with neuromorphic computing. *Nature* **575**, 607–617.

66. Milford MJ, Wyeth GF, Prasser D. 2004 RatSLAM: a hippocampal model for simultaneous localization and mapping. In *IEEE International Conference on Robotics and Automation, 2004. Proceedings. ICRA '04. 2004*, pp. 403–408 Vol.1.

67. Guerguiev J, Lillicrap TP, Richards BA. 2017 Towards deep learning with segregated dendrites. *Elife* **6**. (doi:10.7554/eLife.22901)

68. Goodfellow IJ, Shlens J, Szegedy C. 2014 Explaining and Harnessing Adversarial Examples. *arXiv [stat.ML]*.

69. Brown TB, Mané D, Roy A, Abadi M, Gilmer J. 2017 Adversarial Patch. *arXiv [cs.CV]*

70. Hubel DH, Wiesel TN. 1959 Receptive fields of single neurones in the cat's striate cortex. *J. Physiol.* **148**, 574–591.

71. Lecun Y, Bottou L, Bengio Y, Haffner P. 1998 Gradient-based learning applied to document recognition. *Proc. IEEE* **86**, 2278–2324.

72. Krizhevsky A, Sutskever I, Hinton GE. 2012 ImageNet Classification with Deep Convolutional Neural Networks. In *Advances in Neural Information Processing Systems* (eds F Pereira, CJ Burges, L Bottou, KQ Weinberger), Curran Associates, Inc.

73. Spoerer CJ, Kietzmann TC, Mehrer J, Charest I, Kriegeskorte N. 2020 Recurrent neural networks can explain flexible trading of speed and accuracy in biological vision. *PLoS Comput. Biol.* **16**, e1008215.

74. Cichy RM, Pantazis D, Oliva A. 2014 Resolving human object recognition in space and time. *Nat. Neurosci.* **17**, 455–462.

75. Cichy RM, Khosla A, Pantazis D, Torralba A, Oliva A. 2016 Comparison of deep neural networks to spatio-temporal cortical dynamics of human visual object recognition reveals hierarchical correspondence. *Sci. Rep.* **6**, 1–13.

76. Kriegeskorte N, Wei X-X. 2021 Neural tuning and representational geometry. *Nat. Rev. Neurosci.* **22**, 703–718.

77. Hausmann SB, Vargas AM, Mathis A, Mathis MW. 2021 Measuring and modeling the motor system with machine learning. *Curr. Opin. Neurobiol.* **70**, 11–23.

78. Cao R, Yamins D. 2021 Explanatory models in neuroscience: Part 1 -- taking mechanistic abstraction seriously. *arXiv [q-bio.NC]*.

79. Cao R, Yamins D. 2021 Explanatory models in neuroscience: Part 2 -- constraint-based intelligibility. *arXiv [q-bio.NC]*.

80. Alcorn MA, Li Q, Gong Z, Wang C, Mai L, Ku W-S, Nguyen A. 2018 Strike (with) a Pose: Neural Networks Are Easily Fooled by Strange Poses of Familiar Objects. *arXiv [cs.CV]*.

81. Makhzani A, Shlens J, Jaitly N, Goodfellow I, Frey B. 2015 Adversarial Autoencoders. *arXiv [cs.LG]*.

82. Zhuang C, Yan S, Nayebi A, Schrimpf M, Frank MC, DiCarlo JJ, Yamins DLK. 2021 Unsupervised neural network models of the ventral visual stream. *Proc. Natl. Acad. Sci. U. S. A.* **118**. (doi:10.1073/pnas.2014196118)

83. Tuna OF, Catak FO, Eskil MT. 2022 Uncertainty as a Swiss army knife: new adversarial attack and defense ideas based on epistemic uncertainty. *Complex & Intelligent Systems* (doi:10.1007/s40747-022-00701-0)

84. Reddy MV, Banburski A, Pant N, Poggio T. 2020 Biologically Inspired Mechanisms for Adversarial Robustness. *arXiv [cs.LG]*.

85. Springer JM, Mitchell M, Kenyon GT. 2021 Uncovering Universal Features: How Adversarial Training Improves Adversarial Transferability. *ICML 2021 Workshop on*

86. Guo C, Lee MJ, Leclerc G, Dapello J, Rao Y, Madry A, DiCarlo JJ. 2022 Adversarially trained neural representations may already be as robust as corresponding biological neural representations. *arXiv [q-bio.NC]*.

87. Schrimpf M *et al.* 2020 Brain-Score: Which Artificial Neural Network for Object Recognition is most Brain-Like? *bioRxiv*. , 407007. (doi:10.1101/407007)

88. Rescorla RA, Wagner AR. 1972 A theory of Pavlovian conditioning: Variations on the effectiveness of reinforcement and non-reinforcement. In *Classical conditioning II: Current research and theory* (eds AH Black, WF Prokasy), pp. 64–99. New York: Appleton-Century-Crofts.

89. Harry Klopf A. 1982 *The Hedonistic Neuron: A Theory of Memory, Learning, and Intelligence*. Hemisphere Publishing Corporation.

90. Momennejad I. 2020 Learning Structures: Predictive Representations, Replay, and Generalization.

*Current Opinion in Behavioral Sciences* **32**, 155–166.

91. Schapiro AC, Rogers TT, Cordova NI, Turk-Browne NB, Botvinick MM. 2013 Neural representations of events arise from temporal community structure. *Nat. Neurosci.* **16**, 486–492.

92. Nisioti E, Mahaut M, Oudeyer P-Y, Momennejad I, Moulin-Frier C. 2022 Social Network Structure Shapes Innovation: Experience-sharing in RL with SAPIENS. *arXiv [cs.AI]*.

93. Fedus W, Ramachandran P, Agarwal R, Bengio Y, Larochelle H, Rowland M, Dabney W. 13--18 Jul 2020 Revisiting Fundamentals of Experience Replay. In *Proceedings of the 37th International Conference on Machine Learning* (eds HD Iii, A Singh), pp. 3061–3071. PMLR.

94. Lin L-J. 1992 Self-improving reactive agents based on reinforcement learning, planning and teaching. *Mach. Learn.* **8**, 293–321.

95. McClelland JL. 1998 Complementary learning systems in the brain. A connectionist approach to explicit and implicit cognition and memory. *Ann. N. Y. Acad. Sci.* **843**, 153–169.

96. Pfeiffer BE, Foster DJ. 2013 Hippocampal place-cell sequences depict future paths to remembered goals. *Nature* **497**, 74–79.

97. Deuker L, Olligs J, Fell J, Kranz TA, Mormann F, Montag C, Reuter M, Elger CE, Axmacher N. 2013 Memory Consolidation by Replay of Stimulus-Specific Neural Activity. *J. Neurosci.* **33**, 19373–19383.

98. Lewis PA, Knoblich G, Poe G. 2018 How Memory Replay in Sleep Boosts Creative Problem-Solving. *Trends Cogn. Sci.* **22**, 491–503.

99. Liu Y, Dolan RJ, Kurth-Nelson Z, Behrens TEJ. 2019 Human Replay Spontaneously Reorganizes Experience. *Cell* **178**, 640–652.e14.

100. Momennejad I, Otto AR, Daw ND, Norman KA. 2018 Offline replay supports planning in human reinforcement learning. *Elife*

101. Sutton RS, Szepesvari C, Geramifard A, Bowling MP. 2012 Dyna-Style Planning with Linear Function Approximation and Prioritized Sweeping. *arXiv:1206.3285 [cs]*

102. Schaul T, Quan J, Antonoglou I, Silver D. 2015 Prioritized Experience Replay. *arXiv [cs.LG]*.

103. Horgan D, Quan J, Budden D, Barth-Maron G, Hessel M, van Hasselt H, Silver D. 2018 Distributed Prioritized Experience Replay. *arXiv [cs.LG]*.

104. Barnett SA, Momennejad I. 2022 PARSR: Priority-Adjusted Replay for Successor Representations. In *Reinforcement Learning and Decision Making*,

105. LeCun Y. 2022 A Path Towards Autonomous Machine Intelligence.

106. Adams RA, Huys QJM, Roiser JP. 2016 Computational Psychiatry: towards a mathematically informed understanding of mental illness. *J. Neurol. Neurosurg. Psychiatry* **87**, 53–63.

107. Zorowitz S, Momennejad I, Daw ND. 2020 Anxiety, Avoidance, and Sequential Evaluation. *Computational Psychiatry* **4**, 1–17.

108. Bennett D, Davidson G, Niv Y. 2022 A model of mood as integrated advantage. *Psychol. Rev.* **129**, 513–541.

109. Piloto LS, Weinstein A, Battaglia P, Botvinick M. 2022 Intuitive physics learning in a deep-learning model inspired by developmental psychology. *Nature Human Behaviour*, 1–11.

110. Eleni Nisioti, Katia Jodogne-del Litto, Clément Moulin-Frier. 2021 Grounding an Ecological Theory of Artificial Intelligence in Human Evolution. *NeurIPS 2021 - Conference on Neural Information Processing Systems, Workshop: Ecological Theory of Reinforcement Learning*

111. Coman A, Momennejad I, Drach RD, Geana A. 2016 Mnemonic convergence in social networks: The

112. Momennejad I, Duker A, Coman A. 2019 Bridge ties bind collective memories. *Nat. Commun.* **10**, 1578.

    emergent properties of cognition at a collective level. *Proc. Natl. Acad. Sci. U. S. A.* **113**, 8171–8176.

113. Alonso E, Peter M, Goumard D, Romoff J. 2020 Deep Reinforcement Learning for Navigation in AAA Video Games. *arXiv [cs.LG]*.

114. Nisioti E, Moulin-Frier C. 2020 Grounding Artificial Intelligence in the Origins of Human Behavior. *arXiv [cs.AI]*.

115. Tolman EC. 1948 COGNITIVE MAPS IN RATS AND MEN. *Psychol. Rev.* **55**, 189–208.

116. Rouhani N, Norman KA, Niv Y. 2018 Dissociable effects of surprising rewards on learning and memory. *J. Exp. Psychol. Learn. Mem. Cogn.* **44**, 1430–1443.

117. Zhu Y, Gordon D, Kolve E, Fox D, Fei-Fei L, Gupta A, Mottaghi R, Farhadi A. 2017 Visual Semantic Planning using Deep Successor Representations. *arXiv [cs.CV]*.

118. Sherstan C, Machado MC, Pilarski PM. 2018 Accelerating Learning in Constructive Predictive Frameworks with the Successor Representation. In *2018 IEEE/RSJ International Conference on Intelligent Robots and Systems (IROS)*, pp. 2997–3003.

119. Dayan P. 1993 Improving Generalization for Temporal Difference Learning: The Successor Representation. *Neural Comput.* **5**, 613–624.

120. Machado MC, Rosenbaum C, Guo X, Liu M, Tesauro G, Campbell M. 2017 Eigenoption Discovery through the Deep Successor Representation.

121. Hansen S, Dabney W, Barreto A, Van de Wiele T, Warde-Farley D, Mnih V. 2019 Fast Task Inference with Variational Intrinsic Successor Features. *arXiv [cs.LG]*.

122. de Cothi W *et al.* 2020 Predictive maps in rats and humans for spatial navigation. *bioRxiv*. (doi:10.1101/2020.09.26.314815)

123. Frey M *et al.* 2021 Interpreting wide-band neural activity using convolutional neural networks. *Elife* **10**. (doi:10.7554/eLife.66551)

124. Tampuu A, Matiisen T, Ólafsdóttir HF, Barry C, Vicente R. 2019 Efficient neural decoding of self-location with a deep recurrent network. *PLoS Comput. Biol.* **15**, e1006822.

125. Mavor-Parker AN, Young KA, Barry C, Griffin LD. 2021 Escaping Stochastic Traps with Aleatoric Mapping Agents. *arXiv [cs.LG]*.

126. Behrens TEJ, Muller TH, Whittington JCR, Mark S, Baram AB, Stachenfeld KL, Kurth-Nelson Z. 2018 What Is a Cognitive Map? Organizing Knowledge for Flexible Behavior. *Neuron* **100**, 490–509.

127. Kumaran D, McClelland JL. 2012 Generalization through the recurrent interaction of episodic memories: a model of the hippocampal system. *Psychol. Rev.* **119**, 573–616.

128. Schapiro AC, Turk-Browne NB, Botvinick MM, Norman KA. 2017 Complementary learning systems within the hippocampus: a neural network modelling approach to reconciling episodic memory with statistical learning. *Philos. Trans. R. Soc. Lond. B Biol. Sci.* **372**. (doi:10.1098/rstb.2016.0049)

129. Zhou Z, Singh D, Tandoc MC, Schapiro AC. 2021 Distributed representations for human inference. *bioRxiv*. , 2021.07.29.454337. (doi:10.1101/2021.07.29.454337)

130. Russek EM, Momennejad I, Botvinick MM, Gershman SJ, Daw ND. 2017 Predictive representations can link model-based reinforcement learning to model-free mechanisms. *PLoS Comput. Biol.* **13**, e1005768.

131. Pudhiyidath A, Morton NW, Viveros Duran R, Schapiro AC, Momennejad I, Hinojosa-Rowland DM, Molitor RJ, Preston AR. 2022 Representations of Temporal Community Structure in Hippocampus and Precuneus Predict Inductive Reasoning Decisions. *J. Cogn. Neurosci.* **34**, 1736–1760.

132. Schultz W, Dayan P, Montague PR. 1997 A neural substrate of prediction and reward. *Science* **275**, 1593–1599.

133. Juliani A, Barnett S, Davis B, Sereno M, Momennejad I. 2022 Neuro-Nav: A Library for Neurally-Plausible Reinforcement Learning. *arXiv [cs.NE]*.

134. Ma WJ, Peters B. 2020 A neural network walks into a lab: towards using deep nets as models for human behavior. *arXiv [cs.AI]*.

135. Wilson RC, Collins A. 2019 Ten simple rules for the computational modeling of behavioral data. *https://psyarxiv.com › ...https://psyarxiv.com › ...* (doi:10.31234/osf.io/46mbn)

136. Gershman SJ, Daw ND. 2017 Reinforcement Learning and Episodic Memory in Humans and Animals: An Integrative Framework. *Annu. Rev. Psychol.* **68**, 101–128.

137. Wolpert DH, Macready WG. 1997 No free lunch theorems for optimization. *IEEE Trans. Evol. Comput.* **1**, 67–82.

138. Yang GR, Joglekar MR, Song HF, Newsome WT, Wang X-J. 2019 Task representations in neural networks trained to perform many cognitive tasks. *Nat. Neurosci.* **22**, 297–306.

139. Yang GR, Cole MW, Rajan K. 2019 How to study the neural mechanisms of multiple tasks. *Curr Opin Behav Sci* **29**, 134–143.

140. Márton CD, Zhou S, Rajan K. 2022 Linking task structure and neural network dynamics. *Nat. Neurosci.* **25**, 679–681.

141. Khetarpal K, Riemer M, Rish I, Precup D. 2020 Towards continual reinforcement learning: A review and perspectives. *arXiv preprint arXiv:2012.13490*

142. Degrave J *et al.* 2022 Magnetic control of tokamak plasmas through deep reinforcement learning. *Nature* **602**, 414–419.

143. Van Seijen, Nekoei, Racah. In press. The LoCA regret: a consistent metric to evaluate model-based behavior in reinforcement learning. *Adv. Neural Inf. Process. Syst.*

144. Schrittwieser J *et al.* 2019 Mastering Atari, Go, Chess and Shogi by Planning with a Learned Model. *arXiv [cs.LG]*.

145. Danijar Hafner, Danijar Hafner, Timothy P Lillicrap, Ian FischerShow, James Davidson. 2018 Learning Latent Dynamics for Planning from Pixels. *ICML*

146. Hafner D, Lillicrap TP, Norouzi M, Ba J. 2022 Mastering Atari with Discrete World Models. *https://openreview.net › forumhttps://openreview.net › forum*.

147. Derex M, Boyd R. 2016 Partial connectivity increases cultural accumulation within groups. *Proc. Natl. Acad. Sci. U. S. A.* **113**, 2982–2987.

148. Momennejad I. 2022 Collective minds: social network topology shapes collective cognition. *Philos. Trans. R. Soc. Lond. B Biol. Sci.* **377**, 20200315.

149. Brown TB *et al.* 2020 Language Models are Few-Shot Learners. *arXiv [cs.CL]*.

150. Han S, Pool J, Tran J, Dally WJ. 2015 Learning both weights and connections for efficient neural networks. *arXiv [cs.NE]*.

151. Frankle J, Carbin M. 2018 The Lottery Ticket Hypothesis: Finding Sparse, Trainable Neural Networks. *arXiv [cs.LG]*.

152. Wu C-J *et al.* 2021 Sustainable AI: Environmental implications, challenges and opportunities. *arXiv [cs.LG]*. , 795–813.

153. Siskind JM. 1995 Grounding language in perception. *Artif. Intell. Rev.* **8**, 371–391.

154. Hermann KM *et al.* 2017 Grounded Language Learning in a Simulated 3D World. *arXiv [cs.CL]*.

155. Crawford K. 2021 *The Atlas of AI*. Yale University Press.

156. Binz M, Schulz E. 2022 Using cognitive psychology to understand GPT-3. (doi:10.31234/osf.io/6dfgk)

157. Silver D, Singh S, Precup D, Sutton RS. 2021 Reward is enough. *Artif. Intell.* **299**, 103535.

158. Sutton R. 2019 The Bitter Lesson.

159. Khetarpal K, Ahmed Z, Comanici G, Abel D, Precup D. 2020 What can I do here? A Theory of Affordances in Reinforcement Learning. *arXiv [cs.LG]*.

160. Vaswani A, Shazeer N, Parmar N, Uszkoreit J, Jones L, Gomez AN, Kaiser L, Polosukhin I. 2017 Attention Is All You Need. *arXiv [cs.CL]*.

161. Guiliano E. 1982 *Lewis Carroll: The Complete Illustrated Works*. Gramercy Books.

162. Molchanov P, Tyree S, Karras T, Aila T, Kautz J. 2016 Pruning Convolutional Neural Networks for Resource Efficient Inference. *arXiv [cs.LG]*.

163. Huang Q, Zhou K, You S, Neumann U. 2018 Learning to Prune Filters in Convolutional Neural Networks. In *2018 IEEE Winter Conference on Applications of Computer Vision (WACV)*, pp. 709–718.

164. Zhu MH, Gupta S. 2022 To Prune, or Not to Prune: Exploring the Efficacy of Pruning for Model Compression. *https://openreview.net › forumhttps://openreview.net › forumhttps://openreview.net › pdfhttps://openreview.net › pdf*.

165. Christoff K, Keramatian K, Gordon AM, Smith R, Mädler B. 2009 Prefrontal organization of cognitive control according to levels of abstraction. *Brain Res.* **1286**, 94–105.

166. Shanahan M, Mitchell M. 2022 Abstraction for Deep Reinforcement Learning. *arXiv [cs.LG]*.

167. Momennejad I, Haynes J-D. 2013 Encoding of prospective tasks in the human prefrontal cortex under varying task loads. *J. Neurosci.* **33**, 17342–17349.

168. Russin, O'Reilly, Bengio. In press. Deep learning needs a prefrontal cortex. *Work Bridging AI Cogn*

169. Momennejad I, Haynes JD. 2012 Human anterior prefrontal cortex encodes the 'what'and 'when'of future intentions. *Neuroimage*

170. Hazy TE, Frank MJ, O'reilly RC. 2007 Towards an executive without a homunculus: computational models of the prefrontal cortex/basal ganglia system. *Philos. Trans. R. Soc. Lond. B Biol. Sci.* **362**, 1601–1613.

171. Sakai K. 2008 Task set and prefrontal cortex. *Annu. Rev. Neurosci.* **31**, 219–245.

172. Soltani A, Koechlin E. 2022 Computational models of adaptive behavior and prefrontal cortex. *Neuropsychopharmacology* **47**, 58–71.

173. Hasselmo ME. 2005 A model of prefrontal cortical mechanisms for goal-directed behavior. *J. Cogn. Neurosci.* **17**, 1115–1129.

174. Donoso M, Collins AGE, Koechlin E. 2014 Foundations of human reasoning in the prefrontal cortex. *Science*

175. Badre D, Nee DE. 2018 Frontal Cortex and the Hierarchical Control of Behavior. *Trends Cogn. Sci.* **22**, 170–188.

176. Bunge SA, Wendelken C, Badre D, Wagner AD. 2005 Analogical reasoning and prefrontal cortex: evidence for separable retrieval and integration mechanisms. *Cereb. Cortex* **15**, 239–249.

177. Solé R. 2016 The major synthetic evolutionary transitions. *Philos. Trans. R. Soc. Lond. B Biol. Sci.* **371**.

(doi:10.1098/rstb.2016.0175)

178. Crosby M. In press. Animal AI Olympics. See http://animalaiolympics.com/AAI/ (accessed on 3 July 2022).

179. Keramati M, Gutkin B. 2014 Homeostatic reinforcement learning for integrating reward collection and physiological stability. *Elife* **3**. (doi:10.7554/eLife.04811)

180. Durrant-Whyte H, Bailey T. 2006 Simultaneous localization and mapping: part I. *IEEE Robot. Autom. Mag.* **13**, 99–110.

181. Bailey T, Durrant-Whyte H. 2006 Simultaneous localization and mapping (SLAM): part II. *IEEE Robot. Autom. Mag.* **13**, 108–117.

182. Thrun S. 2008 Simultaneous Localization and Mapping. In *Robotics and Cognitive Approaches to Spatial Mapping* (eds ME Jefferies, W-K Yeap), pp. 13–41. Berlin, Heidelberg: Springer Berlin Heidelberg.